\def\year{2016}
\documentclass[letterpaper]{article}
\usepackage{aaai16}
\usepackage{amsmath}
\usepackage{bbm}
\usepackage{amssymb}
\usepackage{amsthm}
\usepackage{times}
\usepackage{helvet}
\usepackage{courier}
\usepackage{graphicx}
\usepackage{enumitem}
\usepackage{cases}
\usepackage{url}
\usepackage{subfigure}
\usepackage{sistyle}
\usepackage{xcolor} 
\begin{document}
\title{Text Matching as Image Recognition}
\author{Liang Pang$^{*}$, Yanyan Lan$^\dag$, Jiafeng Guo$^\dag$, Jun Xu$^\dag$, Shengxian Wan$^{*}$, \and Xueqi Cheng$^\dag$\\
CAS Key Laboratory of Network Data Science and Technology, \\ Institute of
Computing Technology, Chinese Academy of Sciences, Beijing 100190, China\\
$^*$\{pangliang,wanshengxian\}@software.ict.ac.cn,
$^\dag$\{lanyanyan,guojiafeng,junxu,cxq\}@ict.ac.cn\\
}
\maketitle
\begin{abstract}
\begin{quote}
Matching two texts is a fundamental problem in many natural language processing tasks.
An effective way is to extract meaningful matching patterns from words, phrases, and sentences to produce the matching score.
Inspired by the success of convolutional neural network in image recognition, where neurons can capture many complicated patterns based on the extracted elementary visual patterns such as oriented edges and corners, we propose to model text matching as the problem of image recognition. Firstly, a matching matrix whose entries represent the similarities between words is constructed and viewed as an image. Then a convolutional neural network is utilized to capture rich matching patterns in a layer-by-layer way. We show that by resembling the compositional hierarchies of patterns in image recognition, our model can successfully identify salient signals such as n-gram and n-term matchings. Experimental results demonstrate its superiority against the baselines.
\end{quote}
\end{abstract}

\section{Introduction}
Matching two texts is central to many natural language applications, such as machine translation~\cite{brown1993mathematics}, question and answering~\cite{xue2008retrieval}, paraphrase identification~\cite{socher2011dynamic} and document retrieval~\cite{li2014semantic}.
Given two texts $T_1\!=\!(w_1,w_2,\ldots,w_m)$ and $T_2\!=\!(v_1,v_2,\ldots,v_n)$, the degree of matching is typically measured as a score produced by a scoring function on the representation of each text:
\begin{equation}
	\mathrm{match}(T_1,T_2)=\mathrm{F}\bigl(\Phi(T_1), \Phi(T_2)\bigr),
\end{equation}
where $w_i$ and $v_j$ denotes the $i$-th and $j$-th word in $T_1$ and $T_2$, respectively. $\Phi$ is a function to map each text to a vector, and $\mathrm{F}$ is the scoring function for modeling the interactions between them.

A successful matching algorithm needs to capture the rich interaction structures in the matching process. Taking the task of paraphrase identification for example, given the following two texts:
\begin{description}
  \item[$T_1:$] {\em Down the ages noodles and dumplings were famous Chinese food}.
  \item[$T_2:$] {\em Down the ages dumplings and noodles were popular in China}.
\end{description}
We can see that the interaction structures are of different levels, from words, phrases to sentences.
Firstly, there are many word level matching signals, including identical word matching between
``{\em down}'' in $T_1$ and ``{\em down}'' in $T_2$, and similar word matching between ``{\em famous}'' in $T_1$ and ``{\em popular}''
in $T_2$. These signals compose phrase level matching signals, including n-gram matching between
``{\em down the ages}'' in $T_1$ and ``{\em down the ages}'' in $T_2$, unordered n-term matching between
``{\em noodles and dumplings}'' in $T_1$ and ``{\em dumplings and noodles}'' in $T_2$, and semantic n-term matching between ``{\em were famous Chinese food}'' in $T_1$ and ``{\em were popular in China}'' in $T_2$. They further form sentence level matching signals, which are critical for determining the matching degree of
$T_1$ and $T_2$. How to automatically find and utilize these hierarchical interaction patterns remains a challenging problem.

In image recognition, it has been widely observed that the convolutional neural network (CNN)~\cite{lecun1998gradient,simard2003best} can successfully abstract visual patterns from raw pixels with layer-by-layer composition~\cite{girshick2014rich}. Inspired by this observation, we propose to view text matching as image recognition and use CNN to solve the above problem.
Specifically, we first construct a word level similarity matrix, namely matching matrix, to capture the basic word level matching signals. The matching matrix can be viewed as: 1) a binary image if we define the similarity to be 0-1, indicating whether the two corresponding words are identical; 2) a gray image if we define the similarity to be real valued, which can be achieved by calculating the cosine or inner product based on the word embeddings. Then we apply a convolutional neural network on this matrix. Meaningful matching patterns such as n-gram and n-term can therefore be fully captured within this architecture. We can see that our model takes text matching as a multi-level abstraction of interaction patterns between words, phrases and sentences, with a layer-by-layer architecture, so we name it MatchPyramid.

The experiments on the task of paraphrase identification show that MatchPyramid (with 0-1 matching matrix) outperforms the baselines, by solely leveraging interactions between texts.
While for other tasks such as paper citation matching, where semantic is somehow important, MatchPyramid (with real-valued matching matrix) performs the best by considering both interactions and semantic representations.

Contributions of this paper include: 1) a novel view of text matching as image recognition; 2) the proposal of a new deep architecture based on the matching matrix, which can capture the rich matching patterns at different levels, from words, phrases, to the whole sentences; 3) experimental analysis on different tasks to demonstrate the superior power of the proposed architecture against competitor matching algorithms.

\section{Motivation}
\begin{figure}[!htbp]
	\centering
	\includegraphics[width=0.46\textwidth]{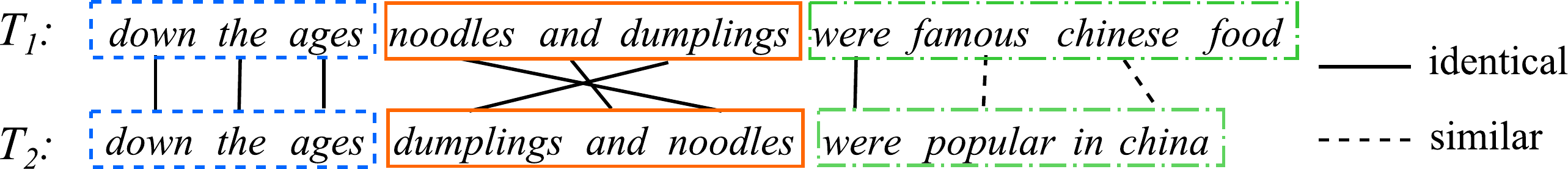}	
\caption{An example of interaction structures in paraphrase identification.}
	\label{Fig.structure}
\end{figure}

It has been widely recognized that making a good matching decision requires to take into account the rich interaction structures in the text matching process, starting from the interactions between words, to various matching patterns in the phrases and the whole sentences. Taking the aforementioned two sentences as an example, the interaction structures are of different levels, as illustrated in Figure~\ref{Fig.structure}.
\begin{figure*}[!htbp]
	\centering
	\includegraphics[width=0.7\textwidth]{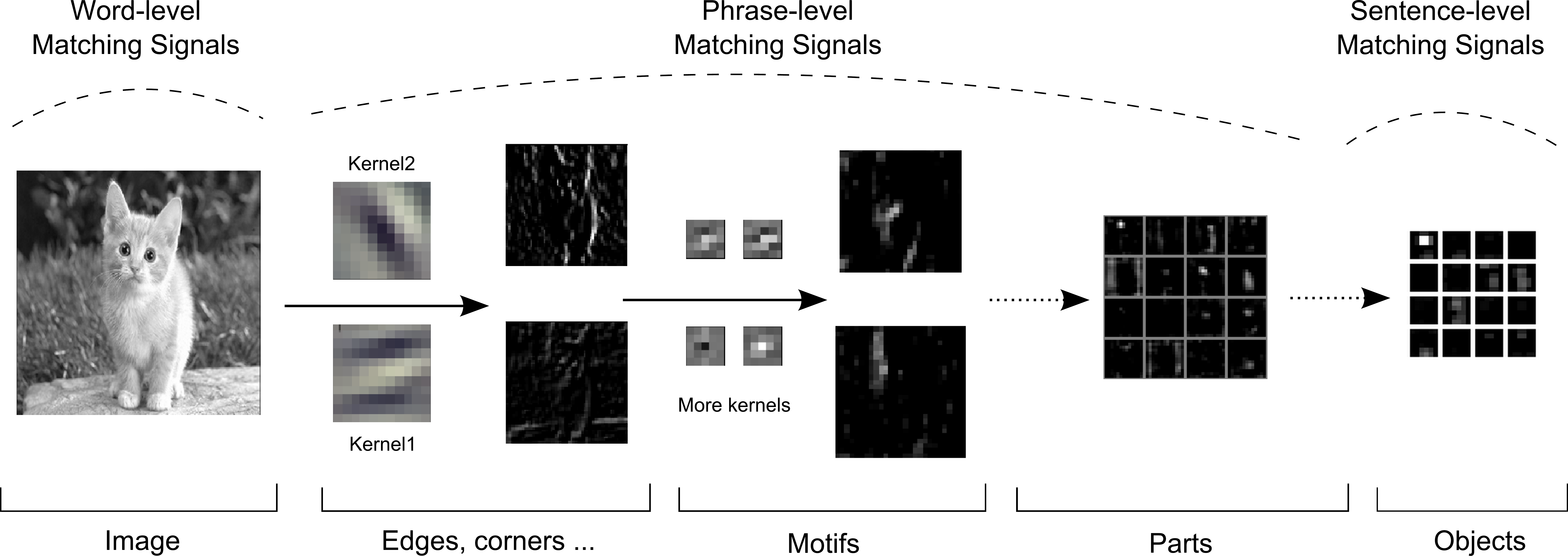}
	\caption{Relationships between text matching and image recognition.}
	\label{Fig.image}
\end{figure*}

\textbf{Word Level Matching Signals} refer to matchings between words in the two texts, including not only identical word matchings, such as ``{\em down}--{\em down}'', ``{\em the}--{\em the}'', ``{\em ages}--{\em ages}'', ``{\em noodles}--{\em noodles}'', ``{\em and}--{\em and}'',``{\em dumplings}--{\em dumplings}'' and ``{\em were}--{\em were}'', but also similar word matchings, such as ``{\em famous}--{\em popular}'' and ``{\em chinese}--{\em china}''.

\textbf{Phrase Level Matching Signals} refer to matchings between phrases, including n-gram and n-term. N-gram matching occurs with n exactly matched successive words,
e.g. ``({\em down the ages})--({\em down the ages})''. While n-term matching allows for order or semantic alternatives, e.g.~``({\em noodles and dumplings})--({\em dumplings and noodles})'', and ``({\em were famous chinese food})--({\em were popular in china})''.

\textbf{Sentence Level Matching Signals} refer to matchings between sentences, which are composed of multiple lower level matching signals, e.g.~the three successive phrase level matchings mentioned above. When we consider matchings between paragraphs that contain multiple sentences, the whole paragraph will be viewed as a long sentence and the same composition strategy would generate paragraph level matching signals.

To sum up, the interaction structures are compositional hierarchies, in which higher level signals are obtained by composing lower level ones. This is similar to image recognition. In an image, raw pixels provide basic units of the image, and each patch may contain some elementary visual features such as oriented edges and corners. Local combinations of edges form motifs, motifs assemble into parts, and parts form objects. We give an example to show the relationships between text matching and image recognition~\cite{jia2014caffe}, as illustrated in Figure~\ref{Fig.image}. In the area of image recognition, CNN has been recognized as one the most successful way to capture different levels of patterns in image~\cite{zeiler2014visualizing}. Therefore, it inspires us to transform text matching to image recognition and employ CNN to solve it. However, the representations of text and image are so different that it remains a challenging problem to perform such transformation.
\section{MatchPyramid}
\begin{figure}[!t]
	\centering
	\includegraphics[width=0.3\textwidth]{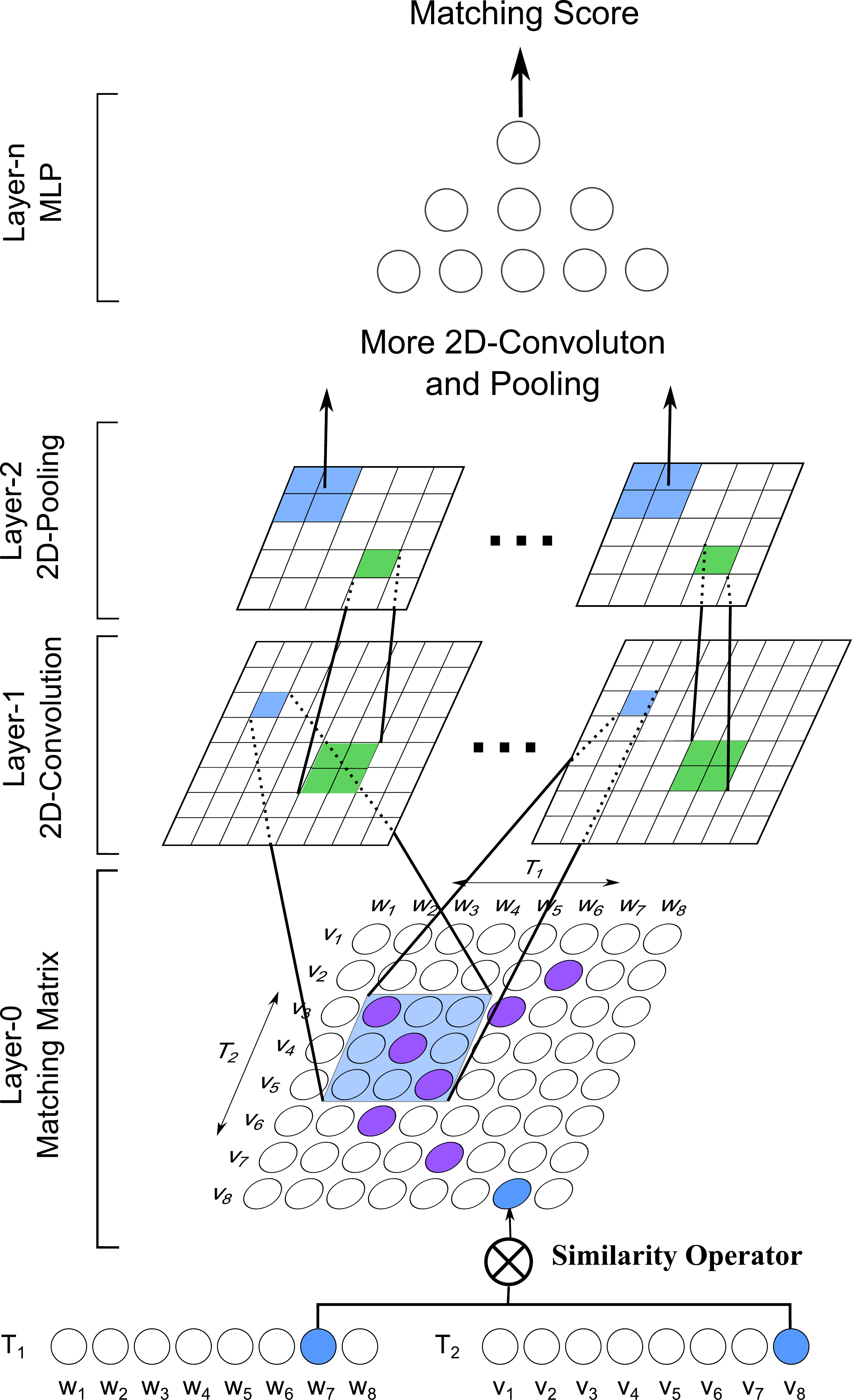}
	\caption{An overview of MatchPyramid on Text Matching.}
	\label{Fig.model}
\end{figure}

In this section we introduce a new deep architecture for text matching, namely MatchPyramid. The main idea comes from modeling text matching as image recognition, by taking the matching matrix as an image, as illustrated in Figure~\ref{Fig.model}.
\subsection{Matching Matrix: Bridging the Gap between Text Matching and Image Recognition}
As discussed before, one challenging problem by modeling text matching as image recognition lies in the different representations of text and image: the former are two 1D (one-dimensional) word sequences while the latter is typically a 2D pixel grid. To address this issue, we represent the input of text matching as a matching matrix $\mathbf{M}$, with each element $\mathbf{M}_{ij}$ standing for the basic interaction, i.e.~similarity between word $w_i$ and $v_j$ (see Eq.~\ref{general_eq}). Here for convenience, $w_i$ and $v_j$ denotes the $i$-th and $j$-th word in two texts respectively, and $\otimes$ stands for a general operator to obtain the similarity.
\begin{equation}
	\label{general_eq}
	\mathbf{M}_{ij} = w_i \otimes v_j.
\end{equation}
In this way, we can view the matching matrix $\mathbf{M}$ as an image, where each entry (i.e.~the similarity between two words) stands for the corresponding pixel value. We can adopt different kinds of $\otimes$ to model the interactions between two words, leading to different kinds of raw images. In this paper, we give three examples as follows.

\textbf{Indicator Function} produces either 1 or 0 to indicate whether two words are identical.
\begin{equation}
\mathbf{M}_{ij}=\mathbb{I}_{\{w_i=v_j\}}=\left\{\begin{aligned}
1, & \qquad\text{if } w_i=v_j\\
0, & \qquad\text{otherwise.}
\end{aligned}\right.
\end{equation}

One limitation of the indicator function is that it cannot capture the semantic matching between similar words. To tackle this problem, we define $\otimes$ based on word embeddings, which will make the matrix more flexible to capture semantic interactions. Given the embedding of each word $\vec{\alpha_i}=\Phi(w_i)$ and $\vec{\beta_j}=\Phi(v_j)$, which can be obtained by recent \texttt{Word2Vec}~\cite{mikolov2013efficient} technique, we introduce the other two operators: cosine and dot product.

\textbf{Cosine} views angles between word vectors as the similarity, and it acts as a soft indicator function.
    \begin{equation}
    	\mathbf{M}_{ij}=\frac{\vec{\alpha_i}^\top \vec{\beta_j}}{\|\vec{\alpha_i}\|\cdot\|\vec{\beta_j}\|},
    \end{equation}
    where $\|\cdot\|$ stands for the norm of a vector, and $\ell_2$ norm is used in this paper.

\begin{figure}[!htbp]
	\centering
	\subfigure[Indicator]{
		\label{Fig.sub.1}
		\includegraphics[width=0.20\textwidth]{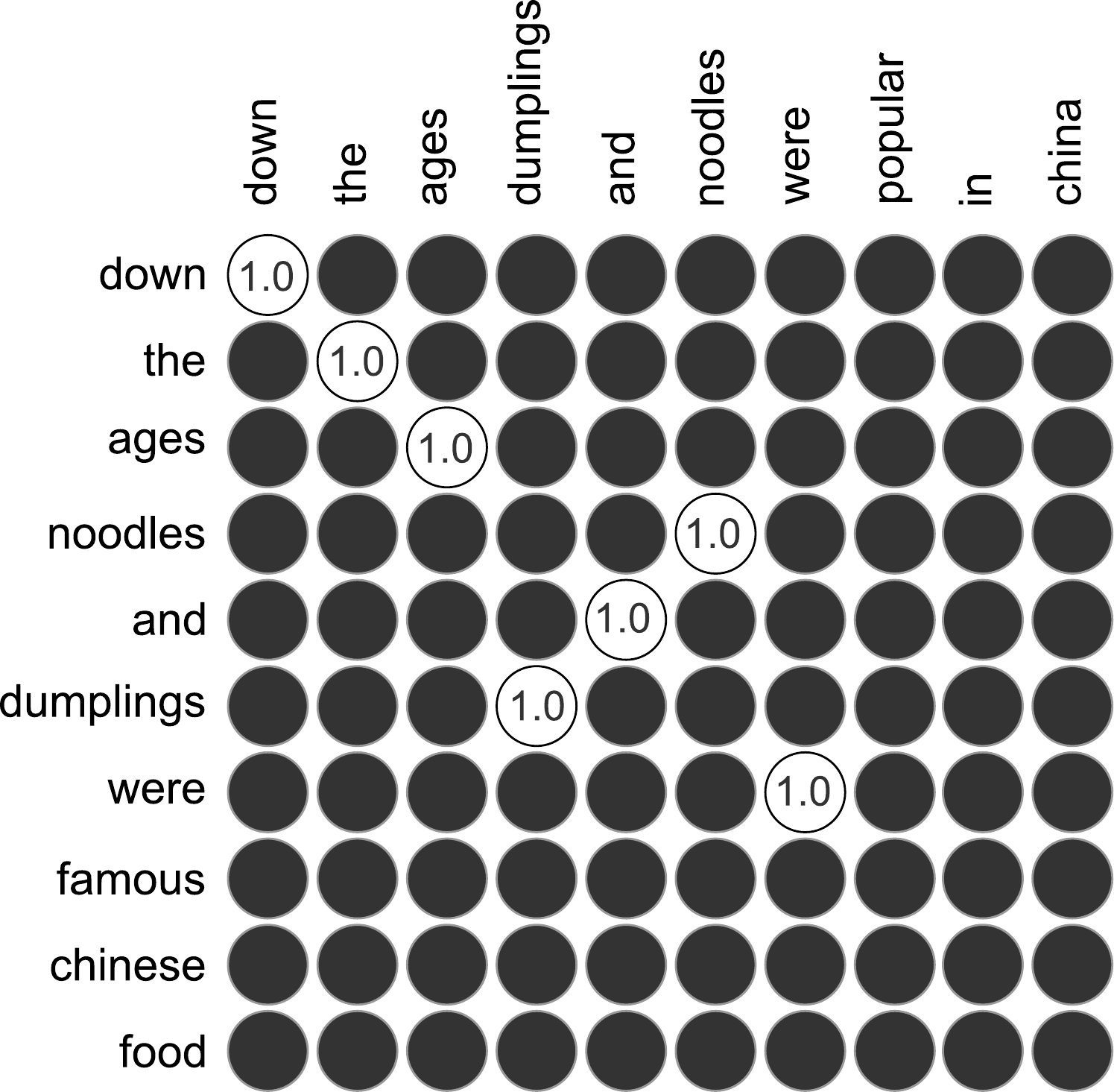}
		}
	\subfigure[Dot Product]{
		\label{Fig.sub.3}
		\includegraphics[width=0.20\textwidth]{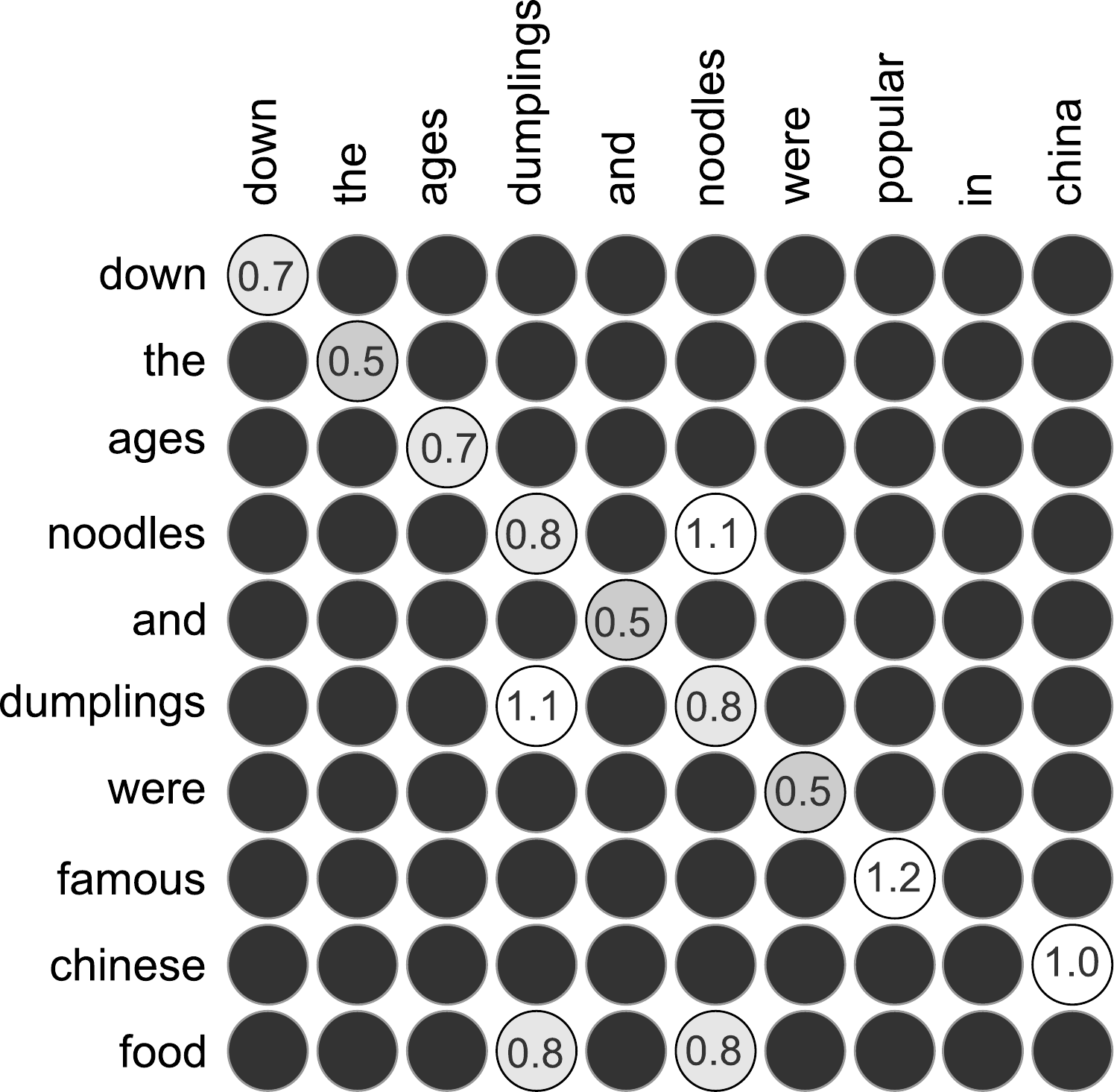}
 	}
 	\caption{Three different matching matrices, where solid circles elements are all valued 0.}
 	\label{Fig.matrix}
\end{figure}

\textbf{Dot Product} further considers the norm of word vectors, as compared to cosine.
    \begin{equation}
        \mathbf{M}_{ij}= \vec{\alpha_i}^\top \vec{\beta_j}.
    \end{equation}
Based on these three different operators, the matching matrices of the given example are shown in Fig~\ref{Fig.matrix}. Obviously we can see that Fig~\ref{Fig.sub.1} corresponds to a binary image, and Fig~\ref{Fig.sub.3} correspond to gray images.

\subsection{Hierarchical Convolution: A Way to Capture Rich Matching Patterns}
The body of MatchPyramid is a typical convolutional neural network, which can extract different levels of matching patterns.
For the first layer of CNN, the $k$-th kernel ${\bf w}^{(1,k)}$ scans over the whole matching matrix ${\bf z}^{(0)}\!=\!\mathbf{M}$ to generate a feature map ${\bf z}^{(1,k)}$:
	\begin{equation}
		{\bf z}_{i,j}^{(1,k)} = \sigma\biggl(\sum_{s=0}^{r_k-1}\sum_{t=0}^{r_k-1} {\bf w}_{s,t}^{(1,k)} \cdot {\bf z}_{i+s,j+t}^{(0)} + b^{(1,k)}\biggr),
	\end{equation}
	where $r_k$ denotes the size of the $k$-th kernel. In this paper we use square kernel, and ReLU~\cite{dahl2013improving} is adopted as the active function $\sigma$.

Dynamic pooling strategy~\cite{socher2011dynamic} is then used to deal with the text length variability. By applying dynamic pooling, we will get fixed-size feature maps:
	\begin{equation}
		{\bf z}_{i,j}^{(2,k)} = \max_{0 \le s < d_k}\max_{0 \le t < d'_k} {\bf z}_{i\cdot d_k+s,j\cdot d'_k+t}^{(1,k)},
	\end{equation}
	where $d_k$ and $d'_k$ denote the width and length of the corresponding pooling kernel, which are determined by the text lengths $n$ and $m$, and output feature map size $n' \times m'$, i.e.~$d_k = \lceil n / n' \rceil, d'_k = \lceil m / m' \rceil$.

After the first convolution and dynamic pooling, we continue to obtain higher level features ${\bf z}^{(l)},l\geq 2$ by further convolution and max-pooling, with general formulations:
	\begin{equation}
        \begin{aligned}
    {\bf z}_{i,j}^{(l+1, k')}&{=}\sigma\biggl(\sum_{k=0}^{c_l-1}\sum_{s=0}^{r_k-1}\sum_{t=0}^{r_k-1} {\bf w}_{s,t}^{(l+1, k')}{\cdot} {\bf z}_{i+s, j+t}^{(l, k)} {+} b^{(l+1, k)}\biggr), \\
        l&=2,4,6,\ldots,
        \end{aligned}
	\end{equation}
	\begin{equation}
        \begin{aligned}
		{\bf z}_{i,j}^{(l+1, k)}&=\max_{0 \le s < d_k}\,\max_{0 \le t < d_k}\, {\bf z}_{i\cdot d_k+s, j\cdot d_k+t}^{(l, k)}, \\
        l&=3,5,7,\ldots,
        \end{aligned}
	\end{equation}
where $c_l$ denote the number of feature maps in the $l$-th layer.

	\begin{figure*}[!htbp]
\centering
        \includegraphics[width=0.7\textwidth]{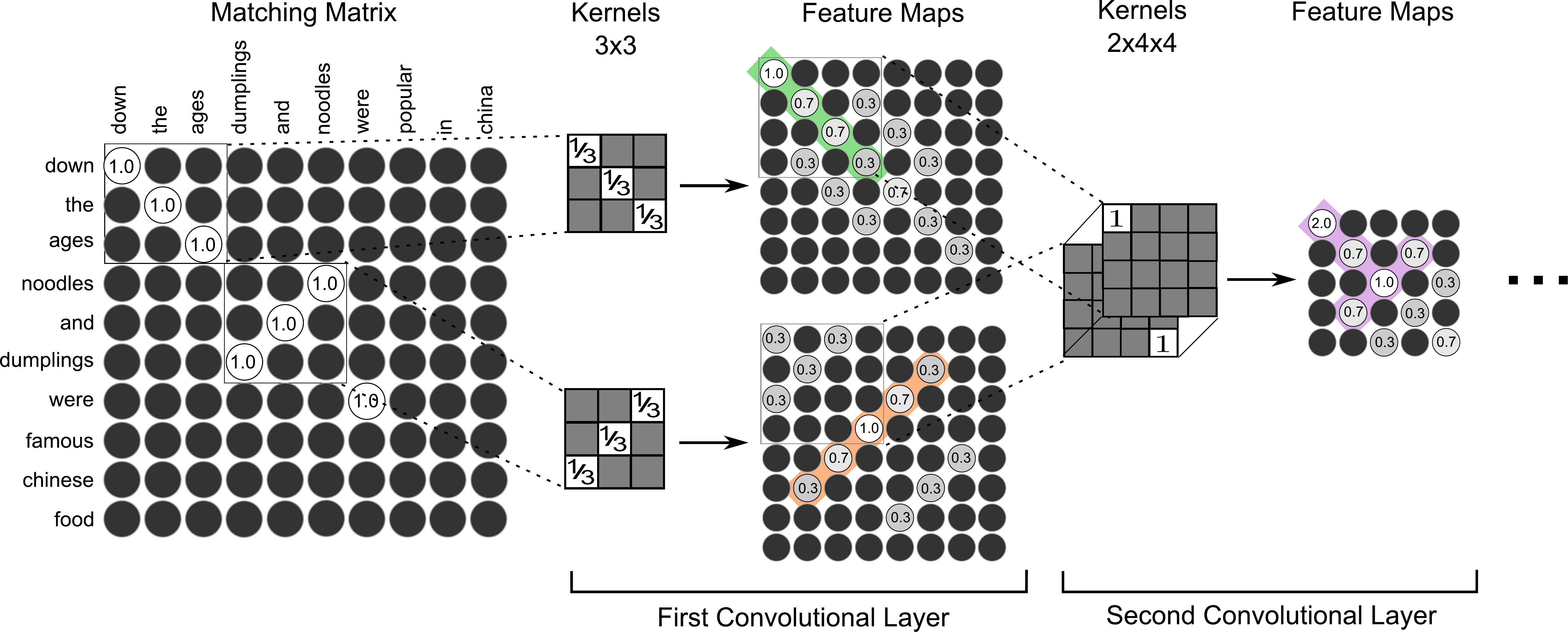}
        \caption{An illustration of Hierarchical Convolution.}
        \label{Fig.kernels}
    \end{figure*}

\textbf{Analysis of Hierarchical Convolution}

Similar to CNN in image recognition where it can make abstractions based on extracted elementary visual patterns such as oriented edges and corners, the hierarchical convolution in MatchPyramid can also capture important phrase level interactions from word level matching and make further compositions. We revisit our example, and show how it works\footnote{Here we take the matching matrix with indicator function as example, similar observations can be obtained for other matching matrix with cosine similarity and dot product.}, as illustrated in Figure~\ref{Fig.kernels}.

(1) With the given two kernels, we can see clearly that the first convolutional layer can capture both n-gram matching signals ``(down the ages)--(down the ages)'' and n-term matching signal ``(noodles and dumplings)--(dumplings and noodles)''. The extracted matching patterns are like edges in image recognition (refer Figure~\ref{Fig.image}).

(2) The following convolutional layers make compositions and form higher level of matching patterns. For example, from the second layer, we can see that a more complicated ``T-type'' pattern captured with the given 3D kernel. It looks like some motif (parts) obtained in image recognition (refer Figure~\ref{Fig.image}).

From the above analysis we can see that MatchPyramid can abstract complicated matching patterns, from phrase to sentence level, by hierarchical convolution.
\subsection{Matching Score and Training}
We use a MLP (Multi-Layer Perception) to produce the final matching score. Take binary classification and two-layer perceptron for example, we will obtain a 2-dimensional matching score vector:
	\begin{equation}
	(s_0,s_1)^\top\!=\!{\bf W}_2\sigma\bigl({\bf W}_1{\bf z} + {\bf b}_1\bigr) + {\bf b}_2,
	\end{equation}
	where $s_0$ and $s_1$ are the matching scores of the corresponding class, ${\bf z}$ is the output of the hierarchical convolution, ${\bf W}_i$ is the weight of the $i$-th MLP layer and $\sigma$ denotes the activation function.

Softmax function is utilized to output the probability of belonging to each class, and cross entropy is used as the objective function for training. Therefore the optimization becomes minimizing:
\begin{equation}
    \begin{aligned}
loss &= -\sum_{i=1}^{N} \Bigl[y^{(i)}\log(p^{(i)}_1)\!+\!(1-y^{(i)})\log(p^{(i)}_0)\Bigr], \\
p_k &= \frac{e^{s_k}}{e^{s_0}+e^{s_1}},\quad k = 0,1,
    \end{aligned}
\end{equation}
where $y^{(i)}$ is the label of the $i$-th training instance. The optimization is relatively straightforward with the standard back-propagation~\cite{williams1986learning}. We apply stochastic gradient descent method Adagrad~\cite{duchi2011adaptive} for the optimization of models. It performs better when we use the mini-batch strategy (32$\sim$50 in size), which can be easily parallelized on single machine with multi-cores. For regularization, we find that some common strategies like early stopping~\cite{giles2001overfitting} and dropout~\cite{hinton2012improving} are enough for our model.
\section{Experiments}
In this section, we conduct experiments on two tasks, i.e. paraphrase identification and paper citation matching, to demonstrate the superiority of MatchPyramid against baselines.
\subsection{Competitor Methods and Experimental Settings}

\textsc{\textbf{AllPositive}}: All of the test data are predicted as positive.

\textsc{\textbf{Tf-Idf}}: \textsc{Tf-Idf}~\cite{salton1983extended} is a widely used method in text mining. In this method, each text is represented as a $|V|$-dimensional vector with each element stands for the \textsc{Tf-Idf} score of the corresponding word in the text, where $|V|$ is the vocabulary size. In this paper, idf score is calculated in the whole dataset. The final matching score is produced by the inner product of the two vectors.

\textsc{\textbf{DSSM/CDSSM}}: Since DSSM~\cite{huang2013learning} and CDSSM~\cite{gao2014modeling,shen2014latent} need large data for training, we directly use the released  models\footnote{http://research.microsoft.com/en-us/downloads/731572aa-98e4-4c50-b99d-ae3f0c9562b9/} (trained on large click-through dataset) on our test data.

\textsc{\textbf{Arc-I/Arc-II}}: We implement \textsc{Arc-I} and \textsc{Arc-II}~\cite{hu2014convolutional} due to there is no publicly available codes, using exactly the same setting as described in the original paper.

There are three versions of MatchPyramid, depending on different methods used for constructing the matching matrices, denoted as \textsc{MP-Ind}, \textsc{MP-Cos}, and \textsc{MP-Dot}, respectively. All these models use two convolutional layers, two max-pooling layers (one of which is a dynamic pooling layer for variable length) and two full connection layers. The number of feature maps is 8 and 16 for the first and second convolutional layer, respectively. While the kernel size is set to be $5 \times 5$ and $3 \times 3$, respectively. Unlike \textsc{Arc-II} which initiates with \texttt{Word2Vec} trained on Wikipedia, we initiate the word vectors in \textsc{MP-Cos} and \textsc{MP-Dot} randomly from a unit ball. Thus our model do not require any external sources.
\subsection{Experiment I: Paraphrase Identification}
	Paraphrase identification aims to determine whether two sentences have the same meaning, a problem considered as a touchstone of natural language understanding. Here we use the benchmark MSRP dataset~\cite{dolan2005automatically}, which contains \num{4076} instances for training and \num{1725} for testing.
    \begin{table}[!htbp]
    \centering
        \caption{Results on MSRP.}
		\label{Table.MSRP}
		\begin{tabular}{l r r}	
			\hline
			Model & Acc.(\%) & $\mathrm{F_1}$(\%) \\
			\hline
			\textsc{AllPositive} & 66.50 & 79.87 \\
			\textsc{Tf-Idf} & 70.31 & 77.62 \\
			\hline
            DSSM & 70.09 & 80.96 \\
            CDSSM & 69.80 & 80.42 \\
			\textsc{Arc-I} & 69.60 & 80.27 \\
			\textsc{Arc-II} & 69.90 & 80.91 \\
			\hline
            \textsc{MP-Ind} & 75.77 & 82.66 \\
            \textsc{MP-Cos} & 75.13 & 82.45 \\
            \textsc{MP-Dot} & \textbf{75.94} & \textbf{83.01} \\
			\hline
		\end{tabular}
	\end{table}	
The experimental results are listed in Table~\ref{Table.MSRP}. We can see that traditional simple model such as \textsc{Tf-Idf} has already achieved a high accuracy of about $70\%$, though it only uses the unigram matching signals. Our methods performs much better than \textsc{Tf-Idf}, which indicates that the complicated matching patterns captured by hierarchical convolution are important to the text matching task. For the comparison with recent deep models, we can see that \textsc{DSSM} performs better than the others (though the improvement is quite limited), and our models (\textsc{MP-Ind}, \textsc{MP-Cos} and \textsc{MP-Dot}) outperform all of them. Though the best performance of our model (75.94\%/83.01\%) is still slightly worse than \textsc{uRAE}~\cite{socher2011dynamic} (76.8\%/83.6\%), \textsc{uRAE} relies heavily on pretraining with an external large dataset annotated with parse tree information. In the future work, we will study how to utilize external data to further improve our models.
	\begin{figure*}[!htbp]
		\centering
		\includegraphics[width=0.7\textwidth]{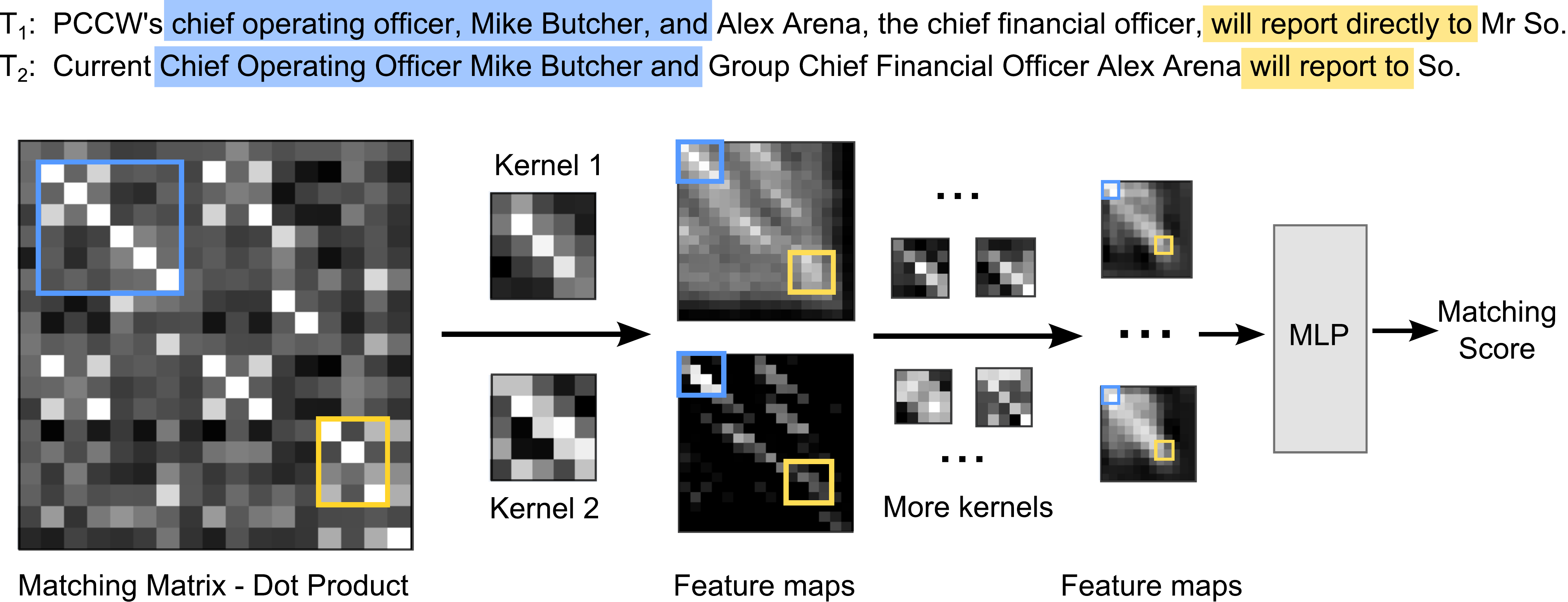}
		\caption{Analysis of the feature maps and kernels in the MatchPyramid Model. The brighter the pixel is, the larger value it has. Better viewed in color.}
		\label{Fig.example}
	\end{figure*}

We also visualize what we have learned in MatchPyramid\footnote{Here we only demonstrate the case of \textsc{MP-dot} due to space limitation. Similar results can be observed with \textsc{MP-Ind} and \textsc{MP-Cos}.}, with expectation that we can gain some insights from the process.	
Specifically, we take a pair of texts as an example (selected from the MSRP dataset), and illustrate the feature maps and kernels in Figure~\ref{Fig.example}. We can see that n-gram and n-term matching, which are emphasized in the blue and yellow color in original texts, are represented as a diagonal sub-matrix emphasized with the blue and yellow rectangles in the matching matrix, respectively. Kernel 1 and kernel 2 are the two kernels learned in the first convolutional layer, which well captures the important n-gram and n-term matching signals respectively. We can see that these patterns are quite similar to the edge extracted by CNN in image recognition (see Figure~\ref{Fig.image}). We also give some more kernels and show some patterns learned in the second convolutional layer. We can see that the latter layer make compositions and keep the useful matching signals until passing it to the MLP classifier. This explains clearly why our model works well: MatchPyramid captures useful matching patterns at different levels, from words, phrase, to sentences, with a similar process in image recognition.		
\subsection{Experiment II: Paper Citation Matching}	
	We evaluate the effectiveness of MatchPyramid with another text matching task called paper citation matching, based on a large academic dataset\footnote{We only use the first 32 words in the abstract.}.
Basically, we are given a set of papers along with their abstracts. A paper and its citations' abstracts then becomes a pair of texts, and defined as a type of matching. One representative example is given as follows:
\begin{itemize}[leftmargin=20pt]
\item[$T_1:$] \emph{this article describes pulsed thermal time of flight ttof flow sensor system as two subsystems pulsed wire system and heat flow system the entire flow sensor is regarded system theoretically as linear.}
\item[$T_2:$] \emph{the authors report on novel linear time invariant lti modeling of flow sensor system based on thermal time of flight tof principle by using pulsed hot wire anemometry thermal he at pulses.}
\end{itemize}
We can see that the matching here should take both lexical and semantic information into consideration. The dataset is collected from a commercial academic website. It contains \num{838908} instances (text pairs) in total, where there are \num{279636} positive (matched) instances and \num{559272} negative (mismatch) instances. The negative instances are randomly sampled papers which have no citation relations. We split the whole dataset into three parts, \num{599196} instances for training, \num{119829} for validation and \num{119883} for testing.			

\begin{table}[!htbp]
    \centering
        \caption{Results on the task of paper citation matching.}
		\label{Table.Paper}
		\begin{tabular}{l r r}
			\hline
			Model & Acc.(\%) & $\mathrm{F_1}$(\%) \\
			\hline
			\textsc{AllPositive} & 33.33 & 50.00\\
			\textsc{Tf-Idf} & 82.63 & 70.21\\
			\hline
            DSSM & 71.97 & 29.88 \\
            CDSSM & 69.84 & 19.97 \\
			\textsc{Arc-I} & 84.51 & 76.79  \\
			\textsc{Arc-II} & 86.48 & 79.57 \\
			\hline
			\textsc{MP-Ind} & 73.76 & 44.71 \\
			\textsc{MP-Cos} & 86.65 & 79.70 \\
			\textsc{MP-Dot} & \textbf{88.73} & \textbf{82.86}  \\
			\hline
		\end{tabular}
	\end{table}

	The results in Table~\ref{Table.Paper} show that \textsc{Tf-Idf} is also a strong baseline on this dataset, which is even better than some deep models such as DSSM and CDSSM.
This may be caused by the large difference between the testing data (paper citation data) and training data (click-through data) used in DSSM and CDSSM.
\textsc{Arc-I} and \textsc{Arc-II} gain a significant improvement over these models, which may benefit much from the large training data.
As for our models, the best performance is still achieved by \textsc{MP-Dot} (88.73\%/82.86\%), which is better than \textsc{Arc-II} (86.84\%/79.57\%).
\textsc{MP-Cos} also gains a better result than \textsc{Arc-II}. The reason of the poor performance of \textsc{MP-Ind} on this task may lie in that the indicator function only captures the exact matching between words, but omits the semantic similarity.

   \begin{table}[!htbp]
		\centering
        \caption{The norm of learned word embeddings on the task of paper citation matching. }
		\label{Table.Word}
		\begin{tabular}{c c c c c c c c c c}
			\hline
            \textbf{Word} & the & with & for & be & are \\
            \hline
            \textbf{Len} & 0.448 & 0.508 & 0.509 & 0.510 & 0.515 \\
            \hline
            \hline
            \textbf{Word}  & robotics & java & snakes & musical & rfid\\
            \hline
            \textbf{Len} & 1.572 & 1.576 & 1.589 & 1.610 & 1.878\\
            \hline
		\end{tabular}
	\end{table}

We further show the reason why \textsc{MP-Dot} performs better than \textsc{MP-Cos} by analyzing the learned word embeddings. Specifically, we pick some words with large and small norm, listed in Table~\ref{Table.Word}. We can see that most words with small norm are indeed useless for matching, while most words with large norm (such as {\em robotics} and {\em java}) are domain terms which play an important role in paper citation matching. By further considering the importance of words, \textsc{MP-Dot} can capture more semantic information than \textsc{MP-Cos} and thus achieve better performance.
\section{Related Work}
Most previous work on text matching tries to find good representations for a single text, and usually use a simple scoring function to obtain the matching results. Examples include Partial Least Square~\cite{wu2013learning}, Canonical Correlation Analysis~\cite{hardoon2003kcca} and some deep models such as DSSM~\cite{huang2013learning}, CDSSM~\cite{gao2014modeling,shen2014latent} and \textsc{Arc-I}~\cite{hu2014convolutional}.

Recently, a brand new approach focusing on modeling the interaction between two sentences has been proposed and gained much attention, examples include \textsc{DeepMatch}~\cite{lu2013deep}, \textsc{uRAE}~\cite{socher2011dynamic} and \textsc{Arc-II}~\cite{hu2014convolutional}. Our model falls into this category, thus we give some detailed discussions on the differences of our model against these methods.

\textsc{\textbf{DeepMatch}} uses topic model to construct the interactions between two texts, and then make different levels of abstractions by a hierarchical architecture based on the relationships between topics. Compared with our matching matrix defined at word level, \textsc{DeepMatch} uses topic information with more rough granularity. Moreover, it relies largely on the quality of learned topic model, and the hierarchies are usually ambiguous since the relationships between topics are not absolute. On the contrary, MatchPyramid clearly models the interactions at different levels, from words, phrases to sentences.

\textsc{\textbf{uRAE}} constructs the interactions between two texts based on the syntactic trees, thus it relies on a predefined compact vectorial representation of text.
Specifically, \textsc{uRAE} first learns the representation of each node on the tree by a auto-encoder, then directly inserts different levels of interaction, such as word, prase and sentence, to a single matrix. Different from that, our MatchPyramid is end-to-end, and captures different levels of interactions in a hierarchical way.

\textsc{\textbf{Arc-II}} and \textsc{Arc-I} are both proposed based on convolutional sentence model DCNN~\cite{kalchbrenner2014convolutional}. Different from \textsc{Arc-I} which defers the interaction of two texts to the end of the process, \textsc{Arc-II} lets them meet early by directly interleaving them to a single representation, and makes abstractions on this basis. Therefore, \textsc{Arc-II} is capturing sentence level interactions directly. However, it is not clear what exactly the interactions are, since they used a sum operation. Our model is also based on a convolutional neural network, but the idea is quite different from that of \textsc{Arc-II}. It is clear that we start from word level matching patterns, and compose to phrase and sentence level matching pattern layer by layer.

\section{Conclusion}
In this paper, we view text matching as image recognition, and propose a new deep architecture, namely MatchPyramid. Our model can automatically capture important matching patterns such as unigram, n-gram and n-term at different levels. Experimental results show that our model can outperform baselines, including some recently proposed deep matching algorithms.

%\balance
\section{Acknowledgments}
This work was funded by 973 Program of China under Grants No.~2014CB340401 and 2012CB316303, 863 Program of China under Grants No.~2014AA015204, the National Natural Science Foundation of China (NSFC) under Grants No.~61472401, 61433014, 61425016, 61425016, and 61203298, Key Research Program of the Chinese Academy of Sciences under Grant No.~KGZD-EW-T03-2, and Youth Innovation Promotion Association CAS under Grants No.~20144310.
\fontsize{9.8pt}{10.8pt}
\selectfont
\bibliographystyle{aaai}
\bibliography{matching_net}

\end{document}